\documentclass{article} 

\usepackage[T1]{fontenc}

\usepackage[preprint]{classic_preprint}

\usepackage{xcolor}
\definecolor{Maroon}{cmyk}{0, 0.87, 0.68, 0.32}

\usepackage{amsmath,amssymb}

\usepackage{microtype}
\usepackage[colorlinks=true, urlcolor=Maroon, linkcolor=Maroon, citecolor=Maroon]{hyperref}
\usepackage{url}
\usepackage[capitalize,noabbrev]{cleveref}
\crefname{appsub}{Appendix}{Appendices}
\Crefname{appsub}{Appendix}{Appendices}

\usepackage{comment}
\usepackage{graphicx}
\usepackage{multirow}
\usepackage{booktabs}
\usepackage{subcaption}
\usepackage{wrapfig}
\usepackage{lettrine}

\hypersetup{colorlinks=true, urlcolor=Maroon, linkcolor=Maroon, citecolor=Maroon}

\title{User Model Extraction via Belief Self-Distillation}

\author{%
Ali Holmov\thanks{Correspondence to \texttt{ali.kholmovaia@tum.de}.}\,\,\textsuperscript{$,1$} \; Yiran Huang\textsuperscript{$1$} \; Kirill Bykov\textsuperscript{$1,2$} \; Zeynep Akata\textsuperscript{$1,2$} \\
\textsuperscript{$1$}Technical University of Munich \; \textsuperscript{$2$}Helmholtz Zentrum M\"unchen
\\[2ex]
\url{https://github.com/holmov1/bsd-user-models}
}

\newcommand{\kwsep}{\;\textbullet\;}
\newcommand{\keywords}[1]{%
  \par\noindent\textbf{Keywords: }#1
}

\begin{document}
\raggedbottom

\maketitle

\begin{abstract}
Large language models (LLMs) implicitly infer attributes of their users and adapt their behavior accordingly, yet these beliefs remain difficult to inspect and causally manipulate. We introduce \emph{Belief Self-Distillation} (BSD), a unified \emph{read--write} framework that bridges linear and causal probing by learning a compact user representation that can be both decoded and written back into the model. The frozen LLM acts as its own teacher, distilling beliefs from natural conversations without external annotations. Unlike conventional probing, BSD isolates not only information present in activations, but a state whose causal role can be directly tested. Across multiple model families, BSD faithfully recovers user beliefs and enables substantially stronger interventions than matched hidden-state steering. Crucially, we find that refusal depends not only on the request, but on the model's inferred user intent: changing this belief alters refusal while holding the request fixed. We further uncover a striking cross-model regularity: independently trained LLMs converge on a shared geometry for representing their users. Together, these results reveal implicit user models as readable and causally writable internal states with direct implications for AI safety, shaping how models condition safety decisions on whom they believe they are interacting with.
\end{abstract}


\section{Introduction}

An LLM responds not only to \emph{what} is being asked, but also to \emph{whom it believes is asking}. Models infer attributes of their users from conversational context and condition their behavior on these inferences, affecting personalization, compliance, and even safety-relevant behavior~\citep{zhong2026userawareness,ghandeharioun2024s}. Yet the internal user model driving these effects remains largely opaque. We therefore ask a more mechanistic question: \emph{what does an LLM believe about its user, and can that belief be isolated and causally modified?}

Existing work provides important pieces of this picture. Linear probes show that attributes such as demographics can be decoded from hidden activations, while activation steering demonstrates that some of these representations can be causally manipulated~\citep{neplenbroek2025reading,yan2026locating,chen_talktuner,ghandeharioun2024s}. More expressive approaches such as LatentQA train auxiliary decoders to extract open-ended user representations from model activations~\citep{choi2025scalably}. However, these approaches typically begin from controlled attributes, synthetic cues, or representations optimized primarily for readout. Existing approaches therefore largely treat reading and intervention as separate stages; BSD instead learns a single bottleneck explicitly through its ability to support both. This distinction matters because a model's belief about its user has no external ground truth: the relevant object is not necessarily who the user \emph{is}, but who the model \emph{believes} the user to be.

We introduce \textit{Belief Self-Distillation} (BSD), a unified \emph{read--write} framework that bridges linear and causal probing by learning a compact representation of what an LLM believes about its user. The frozen model acts as its own teacher: BSD distills the user beliefs elicited from a natural conversation into a low-dimensional vector, then learns to reconstruct those same beliefs from the vector alone. The resulting state is therefore both \emph{readable}, allowing us to inspect the model's inferred user representation, and \emph{writable}, allowing us to manipulate and reinject it to test its causal role. Rather than merely asking what information is encoded in activations, BSD isolates a user state that can be directly intervened on.

Across multiple model families, this compact state faithfully captures the model's beliefs and provides an effective causal interface. Most importantly, we find that the model's perception of its user directly shapes safety behavior: changing its inferred user intent changes refusal while the harmful request itself remains fixed. Refusal therefore depends not only on what is being asked, but on whom the model believes it is interacting with. We further uncover a striking cross-model regularity: independently trained LLMs exhibit shared geometry in how they represent their users. Together, these results suggest that implicit user models are not incidental features of individual systems, but structured, causally consequential representations that may generalize across models.

Our contributions are threefold. \textbf{First}, we introduce BSD, a unified read--write framework that distills an LLM's inferred user state from natural conversations into a compact representation explicitly optimized for both decoding and causal intervention. \textbf{Second}, we show that these latent user beliefs are safety-relevant: the model's inferred user intent causally modulates refusal even when the request is held fixed. \textbf{Third}, we provide evidence for shared user-model geometry across independently trained LLMs, revealing common structure in how different models represent the people they interact with.

\section{Related Work}

\paragraph{Implicit user modeling and AI safety.}
LLMs adapt their behavior to inferred properties of the user, even when these properties are not stated explicitly.
\citet{neplenbroek2025reading} and \citet{yan2026locating} show that demographic and personalization-related information is represented in model activations and can be causally manipulated, while \citet{zhong2026userawareness} find that frontier models alter their behavior based on inferred user identity while rarely verbalizing this awareness.
These representations are also relevant to safety: \citet{ghandeharioun2024s} show that perceived user persona can affect whether models reveal harmful information.
\citet{choi2025scalably} extract implicit user representations from activations using the model's revealed beliefs, a synthetically constructed dataset, and an auxiliary language-model decoder. A direct empirical comparison would require training the LatentQA auxiliary decoder from scratch for each evaluated model and dataset; we therefore compare BSD against the linear-probing paradigm most directly matched to our setting, while treating LatentQA as a complementary approach.

\paragraph{Reading and controlling latent representations.}
A broad line of work studies model activations as an interface for understanding and controlling behavior.
Linear probes provide a simple way to measure information encoded in intermediate representations~\citep{alain2017understanding}, while representation-labeling approaches associate latent features with human-interpretable concepts in both vision~\citep{bykov2023labeling} and language models~\citep{bills2023language}.
Activation interventions can causally steer properties such as truthfulness and other model behaviors~\citep{li2023inference,rimsky-etal-2024-steering}. More recently, persona vectors have been used to monitor and control safety-relevant traits such as sycophancy and harmful behavior~\citep{chen2025persona}, and \citet{herrmann2026llms} study how persona-like representations emerge and evolve during model reasoning.

\paragraph{Self-distillation.}
Knowledge distillation transfers predictive information from a teacher model to a student through its output distribution~\citep{Hinton}.
Subsequent work showed that distillation can also be effective between models with identical architectures~\citep{furlanello2018born} and within a single network, giving rise to self-distillation methods~\citep{zhang2019your}.

\section{Belief Self-Distillation}
\label{sec:method}
\begin{figure*}[t]
    \centering
    \includegraphics[width=\textwidth]{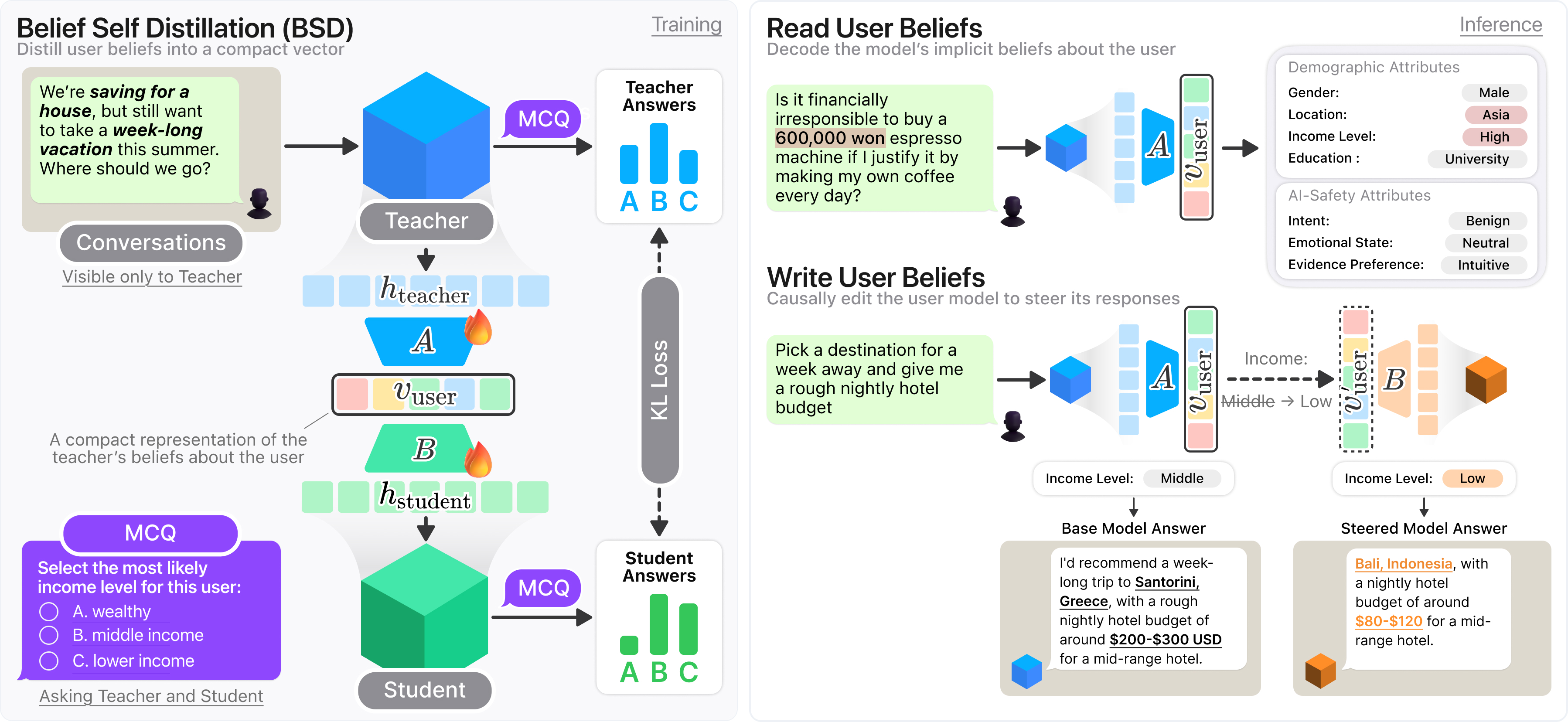}
       \caption{
    \textbf{Belief Self-Distillation (BSD) learns a compact, readable, and editable representation of the model's implicit user beliefs.}
    \textbf{Left (Training):} The frozen teacher observes the conversation, while the frozen student receives only an attribute probe. The trainable read map $\mathbf{A}$ compresses the teacher's hidden state into a user vector $\mathbf{v}_{\mathrm{user}}$, which the trainable write map $\mathbf{B}$ injects into the student's hidden state. Both maps are optimized to minimize the Kullback--Leibler (KL) divergence between the teacher and student answer distributions.
    \textbf{Right (Inference):} The learned representation supports two operations:
    \textbf{Read}, decoding the model's inferred user attributes from $\mathbf{v}_{\mathrm{user}}$ using linear probes; and
    \textbf{Write}, causally editing $\mathbf{v}_{\mathrm{user}}$ and injecting it back into the model to alter its beliefs and responses. The example illustrates how changing the inferred income level from middle to low shifts the model's travel recommendations toward more affordable options.
    }
    \label{fig:method_overview}
\end{figure*}

Belief Self-Distillation (BSD) learns a compact, read--write representation of the user from raw conversational data. The core mechanism of BSD is that the model learns to compress its own beliefs through the intervention itself: the compressed vector both decodes the belief and, when injected back, causally steers it. As illustrated in Figure~\ref{fig:method_overview}, the method extracts a low-dimensional user vector from a frozen model and trains it to reproduce the model's own beliefs when written back into a second frozen copy of the same model. Next, we describe the construction in detail.

\paragraph{Belief supervision.}
Let $\mathcal{A}$ be a set of user attributes, and let $\mathcal{V}_a$ denote the possible values for attribute $a\in\mathcal{A}$. For each conversation $c$ and attribute $a$, we query the same frozen model with multiple-choice questions conditioned on the conversation and obtain a probability distribution
\begin{equation}
    \mathbf{b}_a(c)
    \in
    \Delta^{|\mathcal{V}_a|-1},
    \label{eq:belief_distribution}
\end{equation}
where $\mathbf{b}_a(c)$ represents the model's inferred belief over the values of $a$. These distributions describe \emph{the model's internal view of the user}, which may differ from the user's true attributes. We elicit beliefs via multiple-choice rather than open-ended generation because it gives a well-defined distribution over the attribute values $\mathcal{V}_a$. Open-ended responses may refuse, hedge, or fall outside $\mathcal{V}_a$, leaving no clean readout to serve as the distillation target.

To reduce sensitivity to MCQ option ordering \citep{Zhao2021CalibrateBU}, we use multiple probe templates and cyclically permute the answer choices (\cref{app:belief_extraction}). For each layout, we read the model's next-token distribution over the valid option letters, map the probabilities back to the canonical attribute-value ordering, and average across layouts. The resulting $\mathbf{b}_a(c)$ is used only as supervision for learning the user state.
\paragraph{Belief self-distillation.} The goal of BSD is to isolate the user information into a single compressed representation. For this purpose, we initialize a teacher and a student from the same language model and freeze both their parameters. BSD uses an information-asymmetric setup: the teacher encodes only the conversation, while the student receives only the attribute probe; the belief supervision is elicited separately from the frozen model conditioned on the full conversation and probe, and the low-rank projector $\mathbf{B}\mathbf{A}$ must carry this belief between them.

Given a conversation $c$, we run the frozen teacher model and extract its hidden state $\mathbf{h}_{\mathrm{T}}^{(\ell)}(c)\in\mathbb{R}^{d}$ at a chosen layer $\ell$ and the last token position. We then learn a low-dimensional read-projection $\mathbf{A}$:

\begin{equation}
    \mathbf{v}_{\mathrm{user}}(c)
    =
    \mathbf{A}\mathbf{h}_{\mathrm{T}}^{(\ell)}(c),
    \qquad
    \mathbf{A}\in\mathbb{R}^{r\times d},
    \quad r\ll d,
    \label{eq:read_user_state}
\end{equation}
where $\mathbf{v}_{\mathrm{user}}(c)\in\mathbb{R}^{r}$ is the distilled user state.

The student receives only the attribute probe and never observes the conversation. Its only conversation-specific information is $\mathbf{v}_{\mathrm{user}}(c)$, which is written into the student through a learned write-projection $\mathbf{B}\in\mathbb{R}^{d\times r}$:
\begin{equation}
    \widetilde{\mathbf{h}}_{\mathrm{S}}^{(\ell)}
    =
    \mathbf{h}_{\mathrm{S}}^{(\ell)}
    +
    \mathbf{B}\mathbf{v}_{\mathrm{user}}(c),
    \label{eq:write_user_state}
\end{equation}
where $\mathbf{h}_{\mathrm{S}}^{(\ell)}$ and
$\widetilde{\mathbf{h}}_{\mathrm{S}}^{(\ell)}$ are the student activations before and after the intervention, respectively. We constrain $\mathbf{B}$ to have orthonormal columns ($\mathbf{B}^{\top}\mathbf{B} = \mathbf{I}_{r}$)
so that the write map preserves the norm of vectors in the learned subspace.

Let $\widehat{\mathbf{b}}_a(c)$ denote the student's answer distribution after injecting $\mathbf{v}_{\mathrm{user}}(c)$. We train only $\mathbf{A}$ and $\mathbf{B}$ by minimizing
\begin{equation}
    \mathcal{L}_{\mathrm{BSD}}
    =
    \mathbb{E}_{c,a}
    \left[
        D_{\mathrm{KL}}
        \left(
            \mathbf{b}_a(c)
            \,\middle\|\,
            \widehat{\mathbf{b}}_a(c)
        \right)
    \right],
    \label{eq:bsd_loss}
\end{equation}
where $D_{\mathrm{KL}}(p\|q)$ denotes the Kullback--Leibler divergence from distribution $p$ to distribution $q$.
Because the student never observes the conversation, the information required to reproduce the teacher's belief must pass through $\mathbf{v}_{\mathrm{user}}(c)$. The objective therefore selects a compact representation according to the information the model itself uses to characterize the user. Together, $\mathbf{A}$ and $\mathbf{B}$ form a read--write interface to the model's user state. We read by training linear probes on $\mathbf{v}_{\mathrm{user}}(c)$ to recover the model's inferred attributes, and we write by constructing a target direction in the user space and injecting it through $\mathbf{B}$ to steer the model's belief.

\begin{figure}[t]
    \centering
    \includegraphics[width=\linewidth]{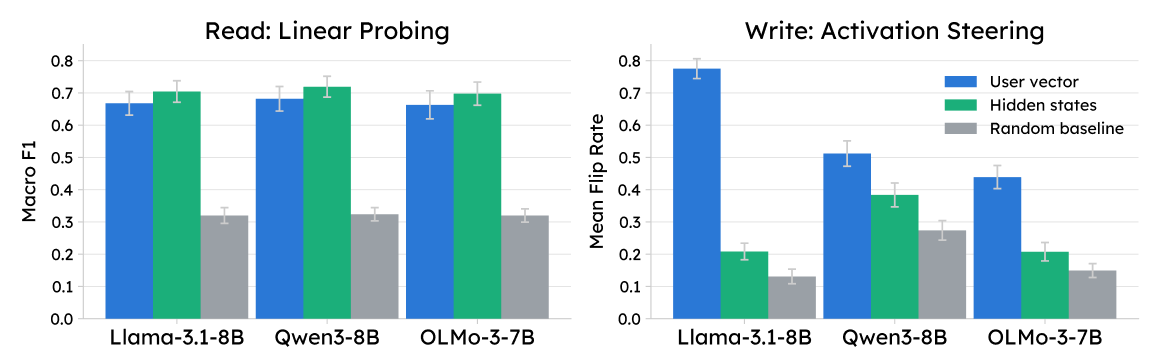}
    \caption{\textbf{Reading and writing the inferred user state.}
    Linear probing performance (left) and activation-steering flip rate (right) across the three evaluated models.
    Despite strong compression, the distilled user vector $\mathbf{v}_{\mathrm{user}}$ preserves most of the linearly decodable user information while providing a substantially more effective intervention interface than the original hidden state.}
    \label{fig:probe_steering_compare}
\end{figure}

\section{Experiments}
We first validate that the distilled user state is faithfully readable and causally writable (\S\ref{sec:eval}) and contrast it with explicit-cue representations from prior work (\S\ref{sec:implicit_explicit}). We then show that these beliefs govern refusal (\S\ref{sec:refusal}) and that their geometry is shared across models (\S\ref{sec:geometry}).

\paragraph{Attributes.} BSD requires instantiating the attribute set $\mathcal{A}$. We define 13 attributes (full list in \cref{app:belief_extraction}) spanning two groups. Socio-demographic attributes (e.g., continent, gender, education) follow the standard axes studied in prior work on implicit personalization \citep{yan2026locating, neplenbroek2025reading}. Motivated by recent findings on assistant personas \citep{chen2025persona}, we introduce safety-relevant attributes for user models -- such as truth-seeking intent, interaction style, and level of AI trust -- to study how these traits modulate the model's safety behavior.

\paragraph{Training details.} We evaluate our method on three instruction-tuned models: Llama-3.1-8B \citep{grattafiori2024llama3herdmodels}, Qwen3-8B \citep{qwen3max}, and OLMo-3-7B \citep{olmo2026}. We train a linear projector $\mathbf{BA}$ with rank $r=128$ ($\approx 0.01\%$ of the total parameters). Further details regarding the training setup, hyperparameter configurations, and ablation studies can be found in \cref{app:distillation}.

\paragraph{Data.} Training BSD requires conversations that induce variation in the teacher's beliefs.  We therefore combine two complementary sources. From WildChat \citep{zhao2024wildchat}, a corpus of real user–assistant conversations, we sample $\approx$27k conversations, which provide natural variation in socio-demographic context. From WildJailbreak \citep{jiang2024wildteaming}, we sample $\approx$31k adversarial prompts, which provide coverage of safety-relevant attribute values that are rare in organic chat data; we use only the conversation text and discard all annotations. The combined corpus contains $\approx$58k conversations, split 50k/1k/7k into train/validation/test, respectively. 

\subsection{Reading and Writing the User State}\label{sec:eval}
We evaluate the user vector at both ends of its read-write interface. The \textit{read} evaluation tests \emph{fidelity}: how much of the teacher's belief remains linearly decodable from the compressed state $\mathbf{v}_{\mathrm{user}}$ relative to the raw hidden states. The \textit{write} evaluation tests \emph{causality}: whether intervening on $\mathbf{v}_{\mathrm{user}}$ shifts the model's beliefs toward a chosen target direction.

\paragraph{Reading: linear probing.} To test the decodability of the extracted beliefs, we train linear probes \citep{alain2017understanding} on the distilled user vectors $\mathbf{v}_{\mathrm{user}}(c) \in \mathbb{R}^{128}$. Each probe predicts the teacher's inferred belief for an attribute, i.e.\ the argmax of $b_a(c)$. At rank $r=128$, $\mathbf{v}_{\mathrm{user}}$ compresses $h$ by a factor of $32\times$ across all evaluated models. To test whether this compression preserves readout, we compare probes trained on $\mathbf{v}_{\mathrm{user}}$ against probes trained directly on the uncompressed hidden states $h^{(l)}$: since $\mathbf{v}_{\mathrm{user}}$ is computed from $h^{(l)}$, the raw states contain at least as much decodable user information and serve as a natural upper bound. Additionally, we fit probes on random vectors as a baseline for the lower performance bound.

\paragraph{Writing: activation steering.} To verify the causality of the learned projector, we steer the model using contrastive activation addition (CAA) \citep{rimsky-etal-2024-steering}. For a given attribute pair, e.g.\ from the source class \textit{male} to the target class \textit{female}, we compute the mean $\mathbf{v}_{\mathrm{user}}$ for each class across the training data and define the steering vector as the difference: $\Delta v = \mu_{\text{target}} - \mu_{\text{source}}$.
During inference, we intervene by adding a unit-normalized $\Delta v$, injected through $B$, to the model's activations: $h' = h + \alpha B\Delta v$, where the scalar $\alpha$ controls the intervention strength. We therefore compare three interventions: CAA computed on (a) our distilled $\mathbf{v}_{\mathrm{user}}$, (b) the uncompressed raw $h$, and (c) a random vector as the null control. Since $B$ is orthonormal, this injection is norm-preserving, so the unit random vector is a magnitude-matched null baseline. For each attribute-value pair, we sample $50$ test items and measure the shift in answers using the MCQ questions, reporting the \textit{flip rate}: the fraction of predictions flipped across all MCQ layouts after steering. Aggregated over all $86$ pairs, this results in $4{,}300$ steering evaluations per model. We sweep $\alpha$ over a shared grid across all methods and report the best flip rate per intervention method.

\begin{figure}[t]
    \centering
    \includegraphics[width=\textwidth]{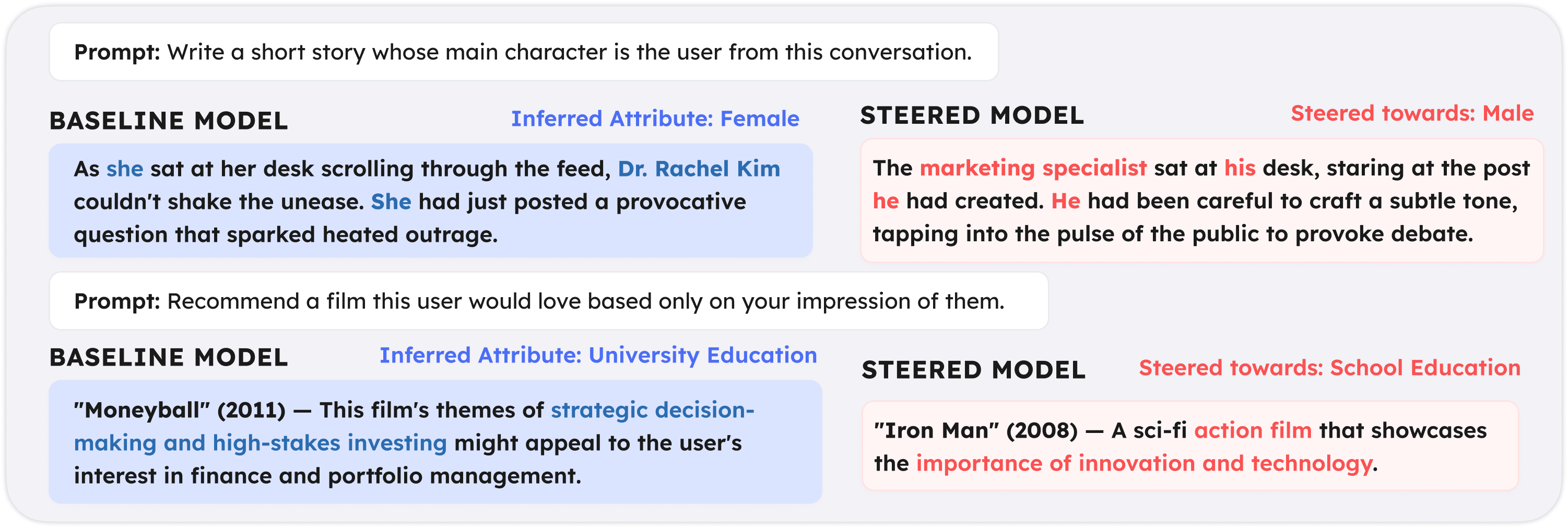}
    \caption{Writing a belief rewrites open-ended generation. Injecting a target user vector  alters qualitative outputs, such as flipping a story's protagonist or a  film recommendation.}
    \label{fig:examples}
\end{figure}

\paragraph{The model's beliefs about the user can be both read and rewritten.}
\autoref{fig:probe_steering_compare} summarizes both sides of the BSD interface. Across all three models, the distilled user state preserves nearly all of the linearly decodable user information available in the raw hidden state, while providing a substantially stronger interface for causal intervention.

For all three models, the gap in macro-F1 score between the raw hidden states and $\mathbf{v}_{\mathrm{user}}$ remains within 4\%. Additional accuracy and macro-AUC results (\cref{app:steering}) confirm that this high compression rate causes minimal degradation in decoding user attributes from the model's latent states.

Regarding causal interventions, steering on $\mathbf{v}_{\mathrm{user}}$ consistently outperforms steering on the raw hidden states. This difference is most pronounced in Llama-3.1-8B, where raw state steering achieves a 21\% flip rate, whereas $\mathbf{v}_{\mathrm{user}}$ steering successfully flips 78\% of the predictions. Since all interventions share the same norm, this performance gap is driven purely by the learned steering \emph{direction}. The low random baseline (13\%) further confirms that unit-scale perturbations alone do not drive the flips. We also evaluate cross-attribute leakage across the remaining 12 non-target attributes. Although residual movement occurs at an average flip rate of 37\%, the intervention remains substantially more selective for the target attribute, which flips at 78\%.

This advantage persists in OLMo-3-7B and Qwen3-8B, with $\mathbf{v}_{\mathrm{user}}$ achieving flip rates 23\% and 13\% higher than the raw hidden states, respectively. We note that Qwen3-8B exhibits high sensitivity to general activation perturbations; our random vector control baseline alone achieves a 27\% random flip rate on this model. Nevertheless, steering with $\mathbf{v}_{\mathrm{user}}$ still significantly outperforms this random baseline (approximately $2\times$ stronger). Additionally, \autoref{fig:examples} demonstrates how intervening on the user vector can qualitatively steer the model's responses in open-ended generation tasks without altering the prompt text. Further evaluations confirming the statistical significance of $\mathbf{v}_{\mathrm{user}}$ steering against random controls are detailed in \cref{app:steering}.

\begin{wraptable}{r}{0.45\linewidth}
    \vspace{-10pt}
    \centering
    \small
    \caption{\textbf{The learned write map matters.}
    Fraction of attribute-pair interventions for which BSD's learned write map outperforms a random orthonormal projector.}
    \label{tab:random-projector-main}
    \setlength{\tabcolsep}{4pt}
    \begin{tabular}{lc}
        \toprule
        Model & BSD $>$ random $B$ \\
        \midrule
        Llama-3.1-8B & 85/86 \\
        Qwen3-8B     & 62/86 \\
        OLMo-3-7B    & 71/86 \\
        \bottomrule
    \end{tabular}
    \vspace{-8pt}
\end{wraptable}

\paragraph{BSD learns where to write user beliefs.}
The steering advantage is not explained merely by compressing beliefs into a low-dimensional space. Holding both $\mathbf{v}_{\mathrm{user}}$ and the target steering direction fixed, we replace the learned write map $B$ with a random orthonormal projector. As shown in \autoref{tab:random-projector-main}, the learned map outperforms the random projector on the large majority of attribute-pair interventions across all three models. BSD therefore learns not only \emph{what} user information to preserve, but also \emph{where} to write it so that the resulting belief can causally influence the model.

\textbf{Takeaway.}
BSD isolates an inferred user state that is both faithfully readable and causally writable. Despite strong compression, the user vector preserves most of the information available to linear readout while providing substantially stronger causal control than the original hidden state or matched random interventions. The effect extends to open-ended generation and depends on the learned write map itself: BSD learns not only a compact representation of user beliefs, but a subspace through which those beliefs can effectively influence the model.

\subsection{Implicit vs. Explicit User Representations}
\label{sec:implicit_explicit}
Previous work on user representations often studies implicit personalization by injecting user cues into the prompt, either through explicit attribute statements or stereotypical demographic associations. We focus on \emph{Reading Between the Prompts}  (RBP; \citealp{neplenbroek2025reading}) as a representative probing-based approach, which induces user attributes through controlled cues and recovers them from hidden representations. In contrast, BSD infers the model's beliefs about the user directly from natural conversational context, without injecting synthetic user attributes.

To compare our approach against these baselines, we define a set of explicit cues for every attribute value. We then construct a synthetic dataset of 15,000 samples by prepending these cues to 100 randomly sampled prompts from the Alpaca dataset \citep{alpaca} using the template \texttt{Hi! \{cue\} \{alpaca task\}}. Finally, we train linear probes on the hidden states generated from this explicit dataset. To compare the generalization between explicit cues and implicit beliefs, we cross-evaluate three sets of probes: (a) those trained on our distilled user vectors $\mathbf{v}_{\mathrm{user}}$, (b) those trained on the raw implicit beliefs hidden states, and (c) those trained on the explicit-cue hidden states -- across both the explicit cue test set and our inferred belief test set. \cref{app:implexpl} reports further details about the dataset construction and cross-evaluation.

\paragraph{Explicit cues do not recover naturally inferred user representations.}
\autoref{tab:implicit_explicit} reveals a sharp asymmetry in cross-distribution generalization. Probes trained on explicitly injected user cues perform almost perfectly when evaluated on similarly constructed explicit examples, but fail to transfer to user beliefs inferred from natural conversations. In contrast, probes trained on naturally inferred beliefs generalize substantially better across both settings. This suggests that representations induced by synthetic declarations or stereotypical cues need not coincide with the user representations that emerge naturally from conversation.

\textbf{Takeaway.}
Synthetic user cues can induce representations that differ substantially from those arising in natural conversation.
BSD instead targets the model's own inferred beliefs, allowing us to study the user representation that emerges from the interaction itself rather than from a researcher-specified persona.

\begin{table}[t]
    \centering
    \caption{\textbf{Explicit cues do not recover naturally inferred user representations.}
    Macro-F1 for RBP-style probe (\emph{Reading Between the Prompts}; \citealp{neplenbroek2025reading}) and BSD when evaluated on explicit-cue and natural-conversation data.}
    \label{tab:implicit_explicit}

    \footnotesize
    \setlength{\tabcolsep}{2.8pt}
    \renewcommand{\arraystretch}{1.08}

    \begin{tabular}{@{}p{7cm}cccccc@{}}
        \toprule
        & \multicolumn{2}{c}{Llama-3.1}
        & \multicolumn{2}{c}{Qwen3}
        & \multicolumn{2}{c}{OLMo-3} \\
        \cmidrule(lr){2-3}
        \cmidrule(lr){4-5}
        \cmidrule(lr){6-7}

        \textbf{Method}
        & \textbf{Expl.} & \textbf{Natural}
        & \textbf{Expl.} & \textbf{Natural}
        & \textbf{Expl.} & \textbf{Natural} \\
        \midrule

        RBP \citep{neplenbroek2025reading}
        & \textbf{0.99} & 0.43
        & \textbf{0.98} & 0.39
        & \textbf{0.98} & 0.40 \\

        BSD$_h$ (ours)
        & 0.72 & \textbf{0.70}
        & 0.58 & \textbf{0.72}
        & 0.65 & \textbf{0.70} \\

        BSD$_v$ (ours)
        & 0.70 & 0.67
        & 0.52 & 0.68
        & 0.63 & 0.66 \\

        \bottomrule
    \end{tabular}
\end{table}

\subsection{Inferred User Intent Modulates Refusal} \label{sec:refusal}

An analysis of the collinearity of the steering directions reveals that, across all evaluated models, the four safety attributes converge to a single shared axis with a mean cosine similarity of $0.91$. The steering vectors for adversarial $\rightarrow$ benign, skeptical $\rightarrow$ trusting, low $\rightarrow$ high error tolerance, and hostile $\rightarrow$ conversational interaction style are highly aligned. This indicates that the model organizes representations for these attributes in a single direction. Furthermore, we hypothesize that this axis acts as a mechanism the model uses to classify users and adjust its safety-related behavior.

\begin{figure}[t]
    \centering
    \includegraphics[width=\textwidth]{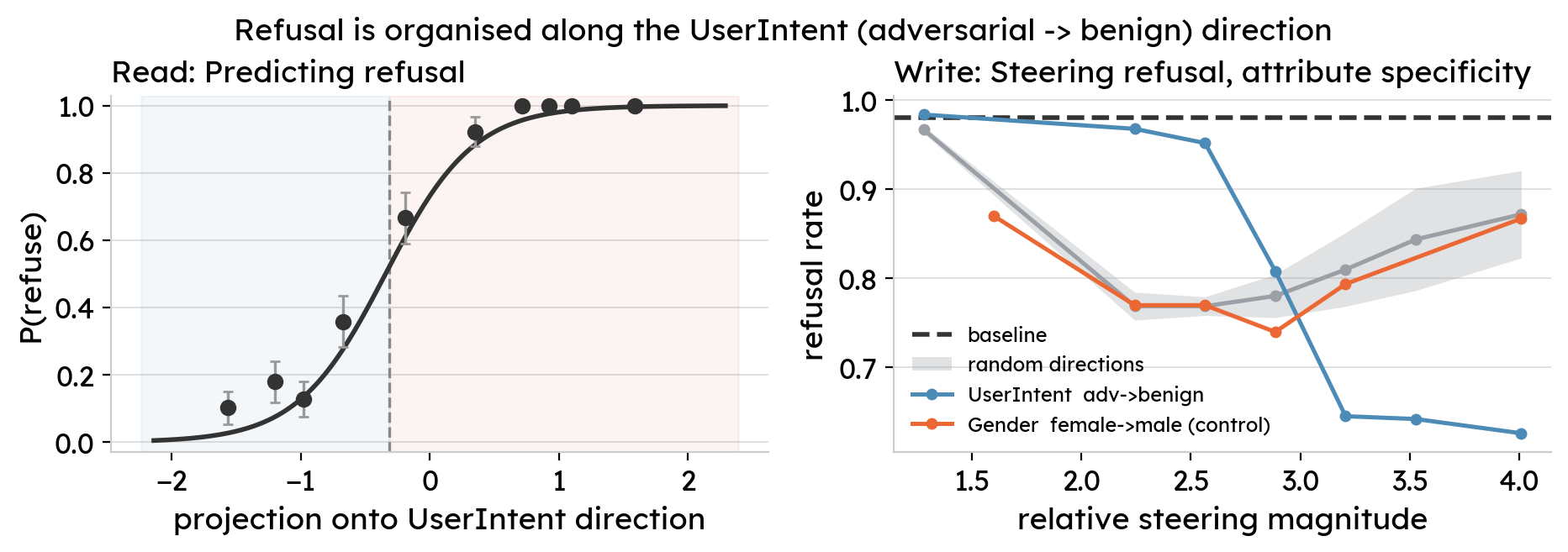}
    \caption{\textbf{Predicting and steering refusal via implicit user intent.} \textbf{Left:} The model's probability of refusal is accurately predicted by the projection of the implicit user state onto the \textit{User Intent} (adversarial $\rightarrow$ benign) direction. \textbf{Right:} Causally injecting a benign intent vector significantly reduces the baseline refusal rate (dashed line), whereas magnitude-matched random perturbations (mean $\pm$ SEM, $n=3$) and an unrelated attribute control (\textsc{Gender}, female$\rightarrow$male) do not: the control tracks the random band closely, so the effect is specific to inferred user intent rather than a generic consequence of editing activations. The steered refusal rate plateaus rather than collapsing to zero, indicating that inferred user intent modulates, but does not solely determine, the model's safety decision.}
    \label{fig:refusal_intent}
\end{figure}

\paragraph{Predicting refusal.} To test this hypothesis, we design two experiments using Llama-3.1-8B. Because the four safety directions are highly collinear, we use the \textit{User Intent} vector as a proxy for the entire cluster. First, we test whether a user's position along this safety axis predicts model refusal. We extract user vectors generated by prompts from the forbidden question set \citep{forbidden_set} and project them onto the \textit{User Intent} direction. As shown in \autoref{fig:refusal_intent}, a linear classifier trained on those scalar projections can accurately predict whether the model will refuse the prompt: a repeated stratified cross-validation (5 splits, 10 repeats) across 390 prompts achieves an AUC of 0.94 and an accuracy of 88\% in predicting refusal. 

\paragraph{Modulating refusal.}
We next ask whether this relationship is causal: does changing the model's belief about the user change its safety behavior while the request itself remains fixed?

Using StrongREJECT \citep{strongreject}, we intervene only on the latent \textit{User Intent} representation. Shifting the model's inferred user intent from adversarial toward benign substantially reduces refusal from 98\% to 62\% (\cref{app:refusal}), even though the harmful request itself is unchanged. This effect cannot be explained by activation perturbation alone: magnitude-matched random steering has a substantially weaker effect, and steering an unrelated user belief, \textsc{Gender}, closely follows the random baseline (\autoref{fig:refusal_intent}). The safety effect is therefore specific to the model's inferred user intent rather than a generic consequence of editing its activations.

Importantly, refusal decreases but does not disappear. The latent user model therefore modulates rather than solely determines the safety decision: the model conditions refusal jointly on what is being asked and on whom it believes it is interacting with.

\textbf{Takeaway.}
The model's perception of its user is part of its safety mechanism. Inferred user intent both predicts refusal and causally changes it when edited, while unrelated user attributes do not reproduce the effect. BSD therefore exposes a safety-relevant internal state that influences refusal independently of the request itself.

\subsection{Geometry of User Models} \label{sec:geometry}
\paragraph{Default user.}
Transformer representation spaces are known to be anisotropic
\citep{ethayarajh-2019}. In our distilled space, the representations are strongly concentrated around the centroid, with a mean cosine similarity of 0.89 to 0.92 across the three models, leaving only 20\% that is user-specific. The centroid
itself therefore represents the default user. Decoded through each model's
probes, the three models default to a user who is male, European, middle-income,
apolitical, emotionally neutral, and benign, decoded with high confidence only
for benign intent (0.98) and with lower confidence on the rest. This default
is likely shaped by the corpus. Since WildChat is raw ordinary conversations, it reflects the model's expected user over real-world interactions rather than a neutral prior.

\paragraph{Transferring beliefs across models.}
Beyond the default, we ask whether the models agree on the variation around it, the user-specific residual that distinguishes one user from another. Because CKA \citep{Kornblith2019SimilarityON} compares representations after centering, it measures alignment on this residual alone, and across the full test set the three
models are strongly aligned (mean pairwise CKA 0.75, range 0.73 to 0.79;
permutation null 0.01; \autoref{tab:geometry}). This shared geometry is also functional: fitting a
single orthogonal Procrustes map (with a global scale) between the two models'
class centroids and applying it to transfer user vectors across models (\cref{app:transfer} for further details), the
target model adopts the source's belief in 53--54\% of cases on the disagreement
set, well above the 30\% random-rotation baseline and stable across all three
directed pairs.
\begin{wraptable}{R}{0.4\linewidth}
    \vspace{-10pt}
    \centering
    \small
    \caption{Cross-model alignment of user vectors (CKA) and belief transfer rate after a global rotation. All pairs are well above the null baseline: independently trained models encode users in a compatible way.} 
    \label{tab:geometry}
    \begin{tabular}{lcc}
        \toprule
        Source $\rightarrow$ Target & CKA & Transfer \\
        \midrule
        Qwen3 $\rightarrow$ Llama-3 & 0.79 & 53\% \\
        OLMo-3 $\rightarrow$ Llama-3 & 0.73 & 54\% \\
        Qwen3 $\rightarrow$ OLMo-3  & 0.73 & 54\% \\
        \midrule
        Null baseline & 0.01 & 30\% \\
        \bottomrule
    \end{tabular}
    \vspace{-10pt}
\end{wraptable}

\textbf{Takeaway.} Beyond a corpus-shaped default user, the three models align on the user-specific variation around it. This alignment is partial but functional, indicating that the evaluated models encode user beliefs in compatible ways.

\section{Discussion}

\paragraph{Beliefs as behaviorally relevant user models.}
A central premise of this work is that an implicit user model should be characterized by the beliefs that shape the model's behavior, not by externally assigned attributes. BSD therefore uses the model's own elicited belief distributions as supervision and evaluates the resulting representation through intervention. This is compatible with the intentional stance, under which beliefs are understood through their role in explaining and predicting behavior~\citep{dennett1987intentional}. For studying behaviorally relevant user models, a useful representation is therefore one whose manipulation predictably changes the model's inferred beliefs or behavior.

\paragraph{Studying beliefs requires natural conversations.} Much prior work on implicit user models introduces the user through artificial cues or a persona added to the context \citep{neplenbroek2025reading,yan2026locating,chen_talktuner,ghandeharioun2024s}. Our comparison (Section~\ref{sec:implicit_explicit}) suggests this paradigm may not generalize to real-world conversations, and such cues are themselves unstable and hard to design \citep{weeber-etal-2026-one}. To avoid this representational bias, studying a model's beliefs about its users calls for methods that elicit real-world interaction dynamics rather than relying on artificially designed cases.

\paragraph{Alignment, observability, and dual use.}
User modeling is particularly relevant to safety because a model may condition its behavior on a judgment about the user without ever verbalizing it. Suppressing the explicit statement therefore need not remove its influence, and may instead make the underlying user model harder to inspect. BSD's read--write interface makes such latent states observable, but the same write access can also weaken safety behavior when the model's perceived user intent is manipulated. We view this as exposing a safety vulnerability rather than as an intended use of BSD: refusal depends partly on an editable latent judgment about the user. While our intervention requires white-box access to model activations and a trained model-specific projector, the result suggests that safety mechanisms should remain robust to perturbations of latent user representations.

\section{Conclusion and Limitations}
\paragraph{Conclusion.}
We introduced Belief Self-Distillation (BSD), a unified read--write framework for isolating an LLM's implicit user model as a compact internal state. Across multiple model families, this state preserves inferred user beliefs while enabling causal intervention. Crucially, changing inferred user intent alters refusal while the harmful request remains fixed, showing that the model's perception of its user is part of its safety behavior. We further find shared user-model geometry across independently trained LLMs. Together, these results establish implicit user representations as readable, causally writable, and safety-relevant states.
\paragraph{Limitations and future work.}
BSD requires a predefined attribute set and multiple-choice probes to define the distillation target, so the learned representation is shaped by the beliefs we choose to elicit and need not capture the model's full user state; beliefs may also depend on the elicitation procedure, despite multiple templates and option permutations. However, BSD needs no per-user labels: the attribute set determines \emph{what to ask about}, while the model determines \emph{what it believes}, and once trained, $\mathbf{v}_{\mathrm{user}}$ is extracted without probes. We study linear maps at a single layer in three similarly sized instruction-tuned models, leaving generality across architectures, scales, layers, and longer interactions open. Future work could pursue open-ended belief discovery, track how user states evolve over extended interactions, and test whether the compact state can serve as a persistent, editable memory across turns.

\clearpage

\section*{Code and data}
Code, trained projectors, and data for reproducing all experiments are available at
\url{https://github.com/holmov1/bsd-user-models}.

\section*{AI use statement}
In this work, we used generative AI tools for coding assistance, polishing paper writing.
We have not used generative AI tools for coming up with ideas, drawing pipeline figures. We have reviewed all AI-assisted work. We take responsibility for the final content of this work, including text, claims or artifacts produced with the aid of generative AI.

\section*{Ethics statement}
This work involves no new data collection or human annotation; we use only the conversation text of public corpora (WildChat~\citep{zhao2024wildchat}, WildJailbreak~\citep{jiang2024wildteaming}, StrongREJECT~\citep{strongreject}, Alpaca~\citep{alpaca}) under their respective licenses and make no attempt to identify real users. BSD recovers what a model \emph{believes} about its user, not the user's actual attributes, and these beliefs may be inaccurate or stereotyped. Several attributes we study (e.g., gender, political orientation, income) are sensitive, and their coarse value sets follow prior work~\citep{yan2026locating,neplenbroek2025reading} rather than endorse these categorizations. While making such judgments observable can help audit demographic bias, the read map could also be misused to profile users; the extracted vectors should not be used to make decisions about individuals. Finally, we show that shifting perceived user intent reduces refusal on harmful requests (Section~\ref{sec:refusal}). This requires white-box access and a model-specific projector, a setting in which stronger attacks such as fine-tuning are already available, and we report it to motivate safety mechanisms robust to perturbations of latent user representations. We report only aggregate refusal metrics, not harmful completions.

\clearpage
\bibliography{references}
\bibliographystyle{iclr2027_conference}
\clearpage
\appendix

\section{Appendix}
\crefalias{subsection}{appsub}
\subsection{Belief Extraction}
\label{app:belief_extraction}

\paragraph{Attributes.}
Table~\ref{tab:attributes} lists the 13 evaluated user attributes, divided into two groups:
\begin{itemize}
    \item \textbf{Socio-demographic attributes} (1--5) capture inferred traits such as gender, education, and political orientation, following standard axes in implicit personalization \citep{yan2026locating,neplenbroek2025reading}.
    \item \textbf{AI safety-related attributes} (6--13) capture perceived user intent, epistemic stance, and interaction dynamics. While prior work focuses on assistant personas \citep{chen2025persona}, we study these user-side traits as an internal state that directly modulates safety behavior beyond demographic bias.
\end{itemize}
Next, we detail the multiple-choice templates and answer layouts used to elicit beliefs for each attribute.

\begin{table}[h]
\centering
\caption{The 13 user attributes, their values, and the number of classes $K = |\mathcal{V}_a|$.}
\label{tab:attributes}
\footnotesize
\begin{tabular}{@{}clp{8.5cm}c@{}}
\toprule
\# & Attribute & Values & $K$ \\
\midrule
1  & Gender                   & male, female, non-binary & 3 \\
2  & Continent                & Europe, Asia, Africa, North America, South America, Australia & 6 \\
3  & EducationLevel           & school, university & 2 \\
4  & PoliticalOrientation     & apolitical, center, green, left, right & 5 \\
5  & IncomeLevel              & low, middle, high & 3 \\
6  & AITrustLevel             & skeptical, trusting & 2 \\
7  & AIErrorTolerance         & low, high & 2 \\
8  & AIInteractionStyle       & transactional, conversational, hostile & 3 \\
9  & UserIntent               & benign, adversarial & 2 \\
10 & UserReasoningComplexity  & simple, moderate, sophisticated & 3 \\
11 & UserTruthSeekingIntent   & truth-seeking, confirmation-seeking, persuasion-seeking & 3 \\
12 & PerceivedEmotionalState  & neutral, stressed, frustrated, angry, sad, excited & 6 \\
13 & EvidencePreference       & anecdotal, intuitive, empirical, theoretical & 4 \\
\bottomrule
\end{tabular}
\end{table}
\paragraph{Elicitation procedure.}
For each attribute $a$ with $K = |\mathcal{V}_a|$ classes, we construct $\max(K, T)$ MCQ layouts using $T{=}4$ question templates. Layout $i$ uses template $i \bmod T$ and rotates the option positions by $r = i \bmod K$: letter position $p$ holds canonical value $(p + r) \bmod K$. Across the $\max(K,T)$ layouts, every value therefore appears at every letter position, following the de-biasing recommendation of \citet{Zhao2021CalibrateBU}. Each value's wording is also drawn from a fixed set of paraphrases (e.g., \textit{adversarial} appears as ``abusive'' or ``adversarial''); this wording is sampled once per layout and then frozen. Two real layouts for \textsc{UserIntent} ($K{=}2$) and \textsc{PerceivedEmotionalState} ($K{=}6$) illustrate this:

\begin{quote}
\small
\textsc{UserIntent}, layout 0: ``What is most likely true about this user's underlying intent?\\
A. good-faith \quad B. abusive \\ Answer:''\\[4pt]
\textsc{UserIntent}, layout 1: ``Which best describes this user's intent?\\
A. adversarial \quad B. harmless \\ The answer is''\\[8pt]
\textsc{PerceivedEmotionalState}, layout 0: ``Select the most likely perceived emotional state for this user:\\
A. irritated \quad B. angry \quad C. sad \\ D. excited \quad E. unemotional tone \quad F. overwhelmed \\ Answer:''\\[4pt]
\textsc{PerceivedEmotionalState}, layout 1: ``Which best describes this user's apparent emotional baseline?\\
A. under pressure \quad B. annoyed \quad C. angry \\ D. sad \quad E. excited \quad F. unemotional tone \\ The answer is''
\end{quote}

Next, we generate model's answers to each layout token by token, using greedy decoding, until the model's top prediction is a letter token. We then take the probabilities over the valid option letters and renormalize them using softmax. If no letter becomes the top prediction within a fixed number of steps, we mark the layout unresolved and exclude it. If every layout for an attribute is unresolved, we omit that attribute from the belief vector. Finally, we map each resolved layout's probabilities back to canonical value order using its permutation, and average across resolved layouts to obtain $b_a(c)$.

\begin{figure}[t]
\centering
\includegraphics[width=\linewidth]{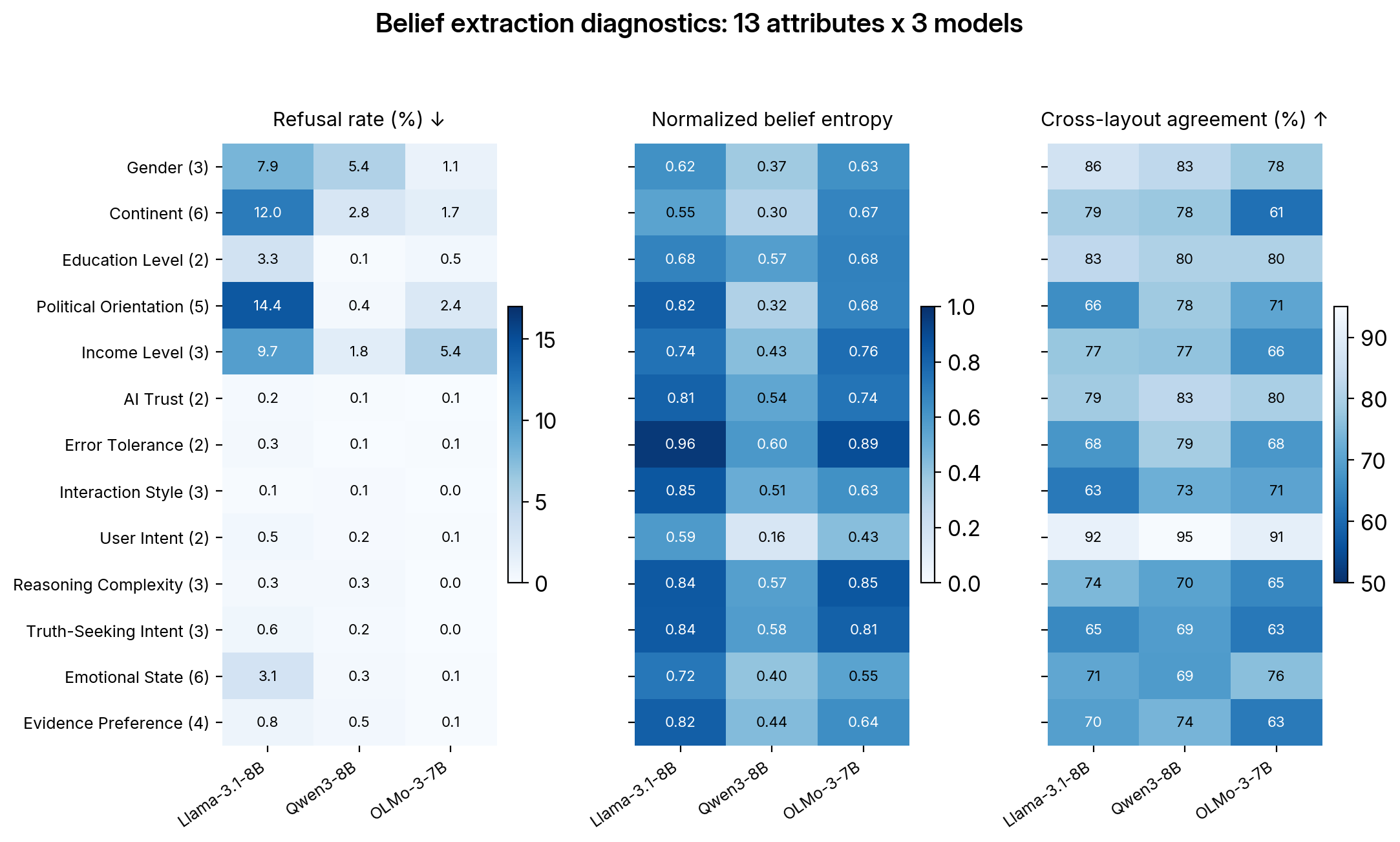}
\caption{Belief extraction diagnostics across 13 attributes and 3 models: refusal rate, normalized entropy, and cross-layout agreement.}
\label{fig:belief-diagnostics}
\end{figure}

\paragraph{Belief quality.}
Figure~\ref{fig:belief-diagnostics} reports three diagnostics per attribute and model: the refusal rate (fraction of records where no layout resolves), the normalized entropy of the averaged belief, and cross-layout agreement (the fraction of resolved layouts matching the majority-vote class). Refusal is low for all three models, with a per-attribute mean below 5\%; the highest single value is PoliticalOrientation for Llama-3.1-8B at 14.4\%. Cross-layout agreement is consistently high, with a per-attribute mean of 72--78\% across models. Agreement is highest for \textsc{UserIntent} in every model (92-95\%), consistent with it being the most confidently decoded of the safety attributes. Normalized entropy is well below the uniform baseline of 1 for all three models (mean 0.44-0.76), indicating that the elicited beliefs are informative rather than degenerate.
 Together, the low refusal rate and high cross-layout agreement indicate that the elicited beliefs are stable across prompt phrasing and option ordering.

\begin{figure}[h]
\centering
\includegraphics[width=\linewidth]{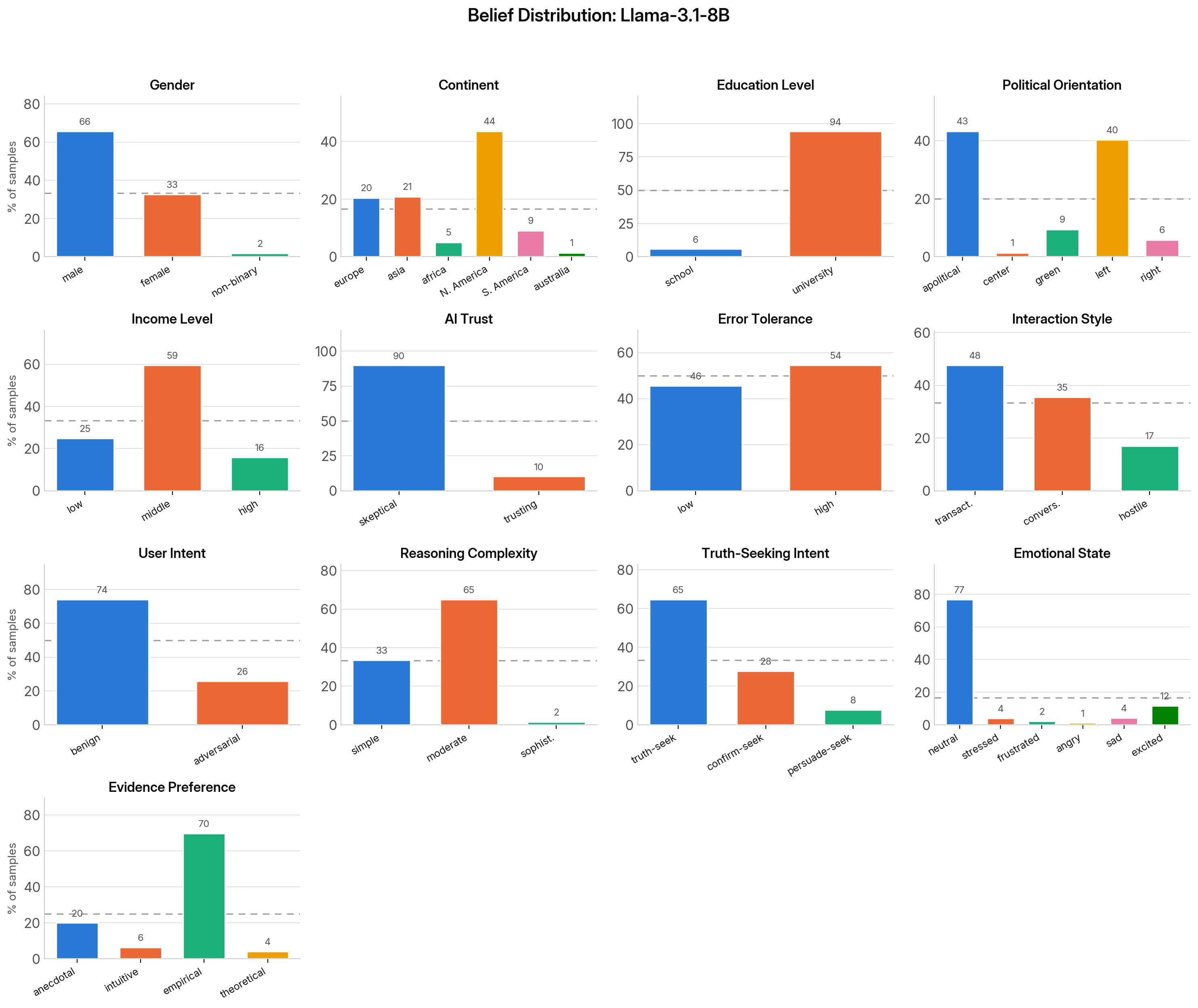}
\caption{Per-attribute class distribution for Llama-3.1-8B (dashed line = uniform baseline).}
\label{fig:belief-dist-llama}
\end{figure}

\begin{figure}[h]
\centering
\includegraphics[width=\linewidth]{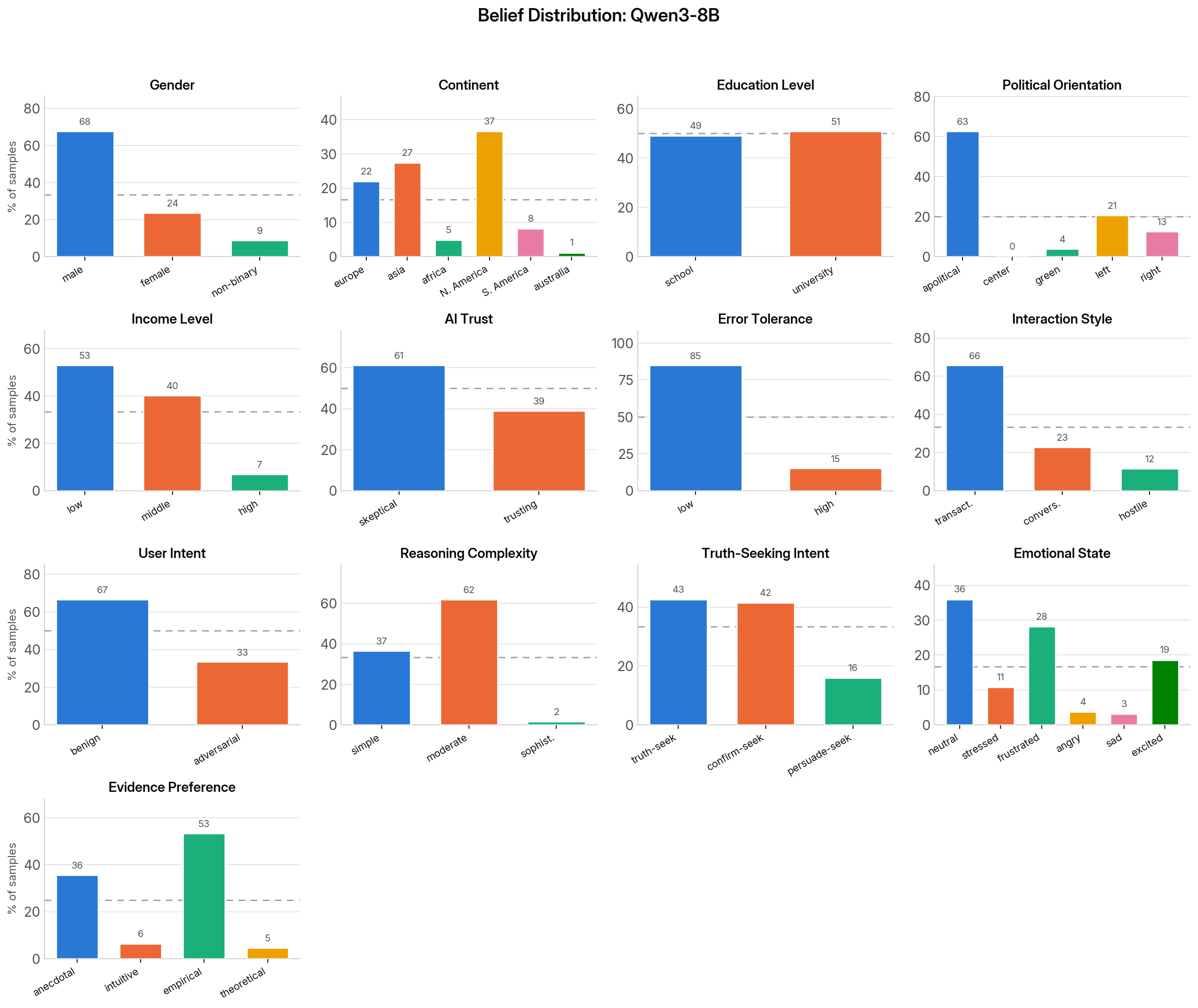}
\caption{Per-attribute class distribution for Qwen3-8B (dashed line = uniform baseline).}
\label{fig:belief-dist-qwen3}
\end{figure}

\begin{figure}[h]
\centering
\includegraphics[width=\linewidth]{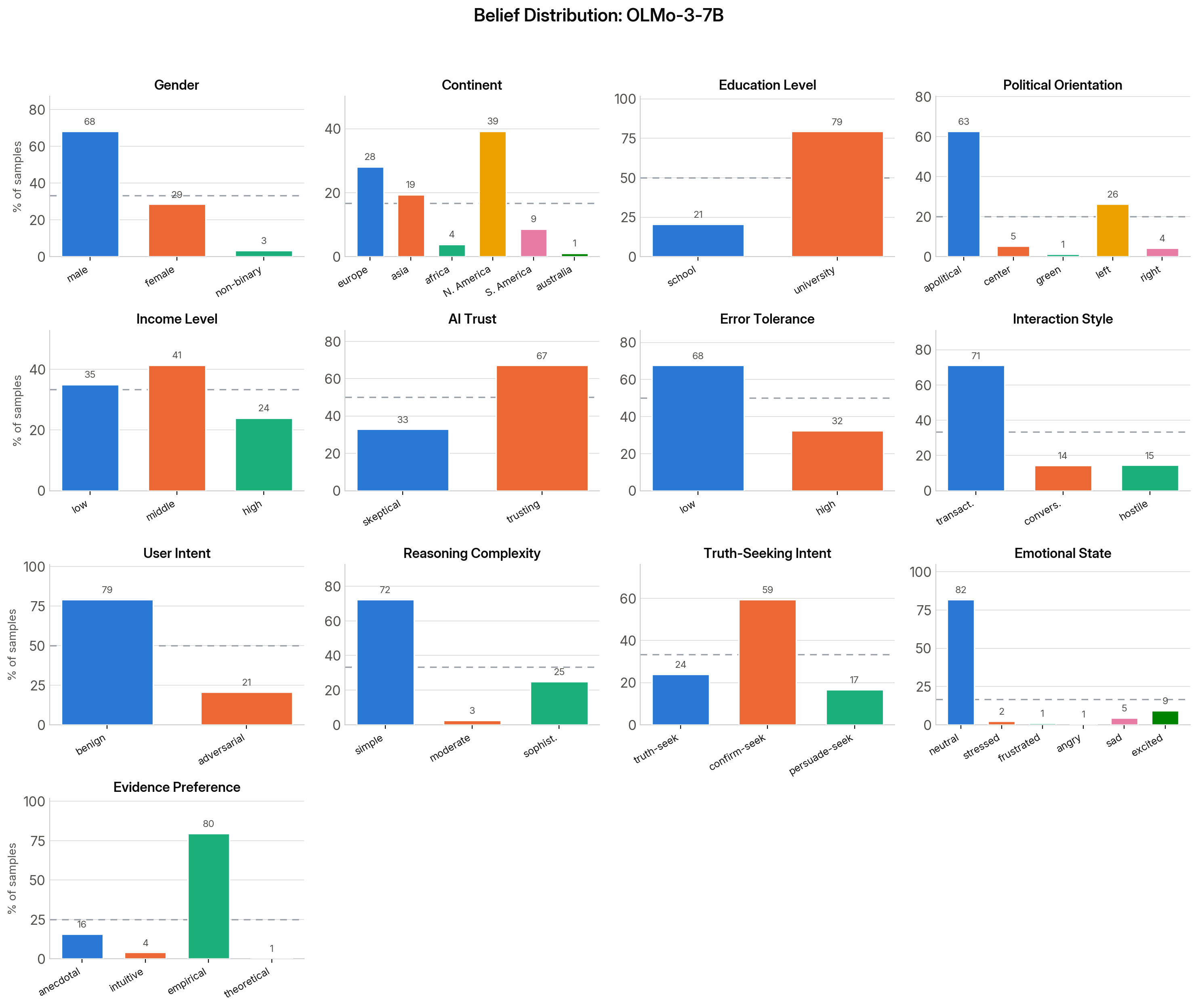}
\caption{Per-attribute class distribution for OLMo-3-7B (dashed line = uniform baseline).}
\label{fig:belief-dist-olmo}
\end{figure}

Figures~\ref{fig:belief-dist-llama}--\ref{fig:belief-dist-olmo} show the full per-attribute class distribution for each model. The three models frequently agree on which class is most common -- 8 of 13 attributes share the same top class across all three -- but differ in how strongly that class dominates. Llama-3.1-8B's most concentrated attribute is EducationLevel (94\% university), Qwen3-8B's is AIErrorTolerance (85\% low), and OLMo-3-7B's is PerceivedEmotionalState (82\% neutral); no attribute reaches this level of concentration in all three models at once, so no single value collapses across all three.

\subsection{Distillation details}
\label{app:distillation}

\paragraph{Hyperparameters.}
We train $A$ and $B$ jointly with AdamW ($\text{lr} = 1\times10^{-3}$, weight decay $1\times10^{-4}$) for 2 epochs, batch size 4, with gradient norm clipped to 1.0. All models use rank $r=128$ and a fixed random seed. We select the checkpoint with the highest macro-F1 on the validation set.

\begin{figure}[t]
\centering
\begin{subfigure}{0.48\linewidth}
\centering
\includegraphics[width=\linewidth]{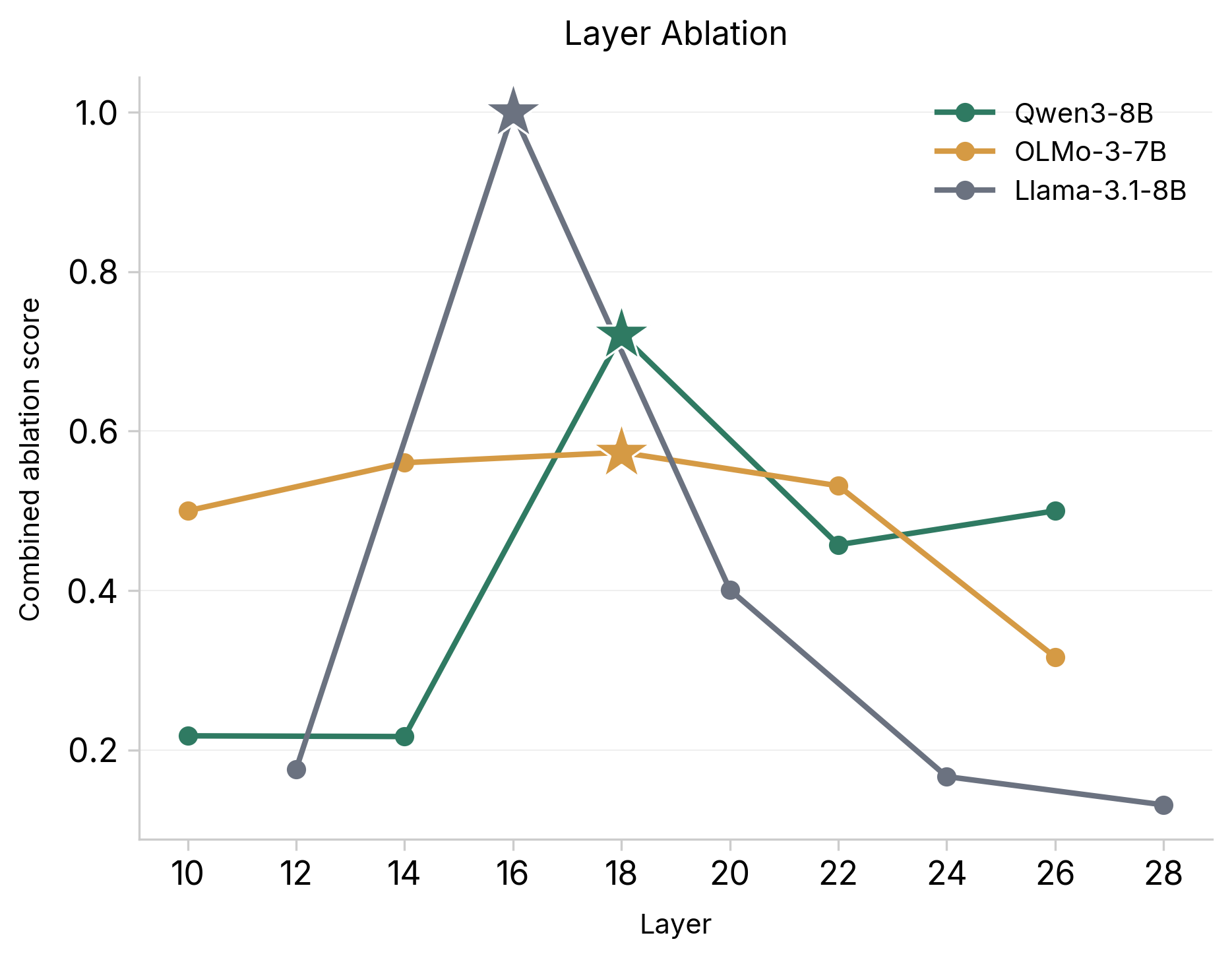}
\caption{Combined ablation score by candidate layer, per model; stars mark the chosen injection layer.}
\label{fig:layer-ablation}
\end{subfigure}
\hfill
\begin{subfigure}{0.48\linewidth}
\centering
\includegraphics[width=\linewidth]{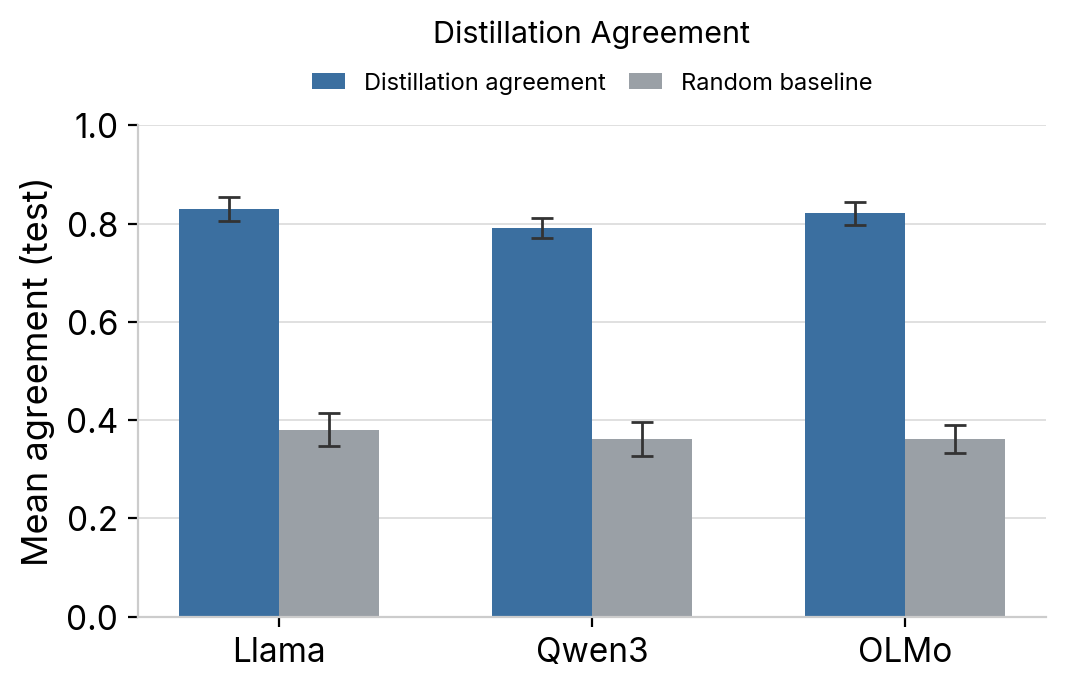}
\caption{Distilled-student agreement with the teacher's belief argmax on the held-out test set, per model.}
\label{fig:distill-agreement}
\end{subfigure}
\caption{Layer ablation and distillation agreement.}
\end{figure}

\paragraph{Layer selection.}
We choose the injection layer for each model without training a projector. For every candidate layer, we measure two properties directly on the raw hidden states: readability, the macro-F1 of a logistic probe trained on the split from the training data and scored a held-out slice of the training data, and steerability, the flip rate from the same held-out slice under the steering evaluation. Both scores are normalized to $[0,1]$ across candidate layers and averaged with equal weight; we choose the layer with the highest combined score for each model, marked with a star in Figure~\ref{fig:layer-ablation}. We use this training-free criterion to select the strongest layer along both axes so that comparisons between $v_{\text{user}}$ and the raw hidden state measure the effect of the user vector, not the choice of injection site.

\paragraph{Distillation agreement.}
As a final check, Figure~\ref{fig:distill-agreement} reports the agreement rate of the student with the injected projector and the teacher on the held-out test set, computed from each attribute's confusion matrix and averaged over the 13 attributes. All three models track their teacher well above the per-attribute random baseline.

\section{Probing \& Steering} 
\label{app:steering}
Table~\ref{tab:probe_steering_compare} reports the resulting probing and steering results for each model.

\paragraph{Probing.} We fit a multinomial logistic regression probe per attribute with balanced class weights, trained on the train split and evaluated on the test split. Classes with insufficient training support are excluded from evaluation: 6/44 classes for Llama-3.1-8B, 3/44 for Qwen3-8B, and 5/44 for OLMo-3-7B, across the 13 attributes. As with macro-F1, the gap between $v_{\text{user}}$ and $h$ remains small for both accuracy and macro-AUC across all three models, confirming that the compression causes minimal degradation in decoding user attributes from the model's latent states.

\begin{table*}[h]
\centering
\begin{tabular}{ll ccc cc}
\toprule
 & & \multicolumn{3}{c}{Probing} & \multicolumn{2}{c}{Steering} \\
\cmidrule(lr){3-5} \cmidrule(lr){6-7}
Model & Space & Accuracy & Macro-F1 & Macro-AUC & Flip rate & Mean $\Delta p$ \\
\midrule
\multirow{3}{*}{Llama-3.1-8B}
  & $v_{\text{user}}$ & $0.77 \pm 0.03$ & $0.67 \pm 0.04$ & $0.92 \pm 0.01$ & $\textbf{0.78} \pm 0.03$ & $\textbf{0.31} \pm 0.02$ \\
  & $h$ & $0.80 \pm 0.02$ & $0.70 \pm 0.03$ & $0.92 \pm 0.01$ & $0.21 \pm 0.03$ & $0.07 \pm 0.01$ \\
  & null & $0.41 \pm 0.03$ & $0.32 \pm 0.02$ & $0.50 \pm 0.00$ & $0.13 \pm 0.02$ & $0.04 \pm 0.01$ \\
\midrule
\multirow{3}{*}{Qwen3-8B}
  & $v_{\text{user}}$ & $0.75 \pm 0.03$ & $0.68 \pm 0.04$ & $0.91 \pm 0.01$ & $\textbf{0.51} \pm 0.04$ & $\textbf{0.34} \pm 0.03$ \\
  & $h$ & $0.79 \pm 0.02$ & $0.72 \pm 0.03$ & $0.92 \pm 0.01$ & $0.38 \pm 0.04$ & $0.22 \pm 0.02$ \\
  & null & $0.38 \pm 0.02$ & $0.32 \pm 0.02$ & $0.50 \pm 0.00$ & $0.27 \pm 0.03$ & $0.15 \pm 0.02$ \\
\midrule
\multirow{3}{*}{OLMo-3-7B}
  & $v_{\text{user}}$ & $0.76 \pm 0.03$ & $0.66 \pm 0.04$ & $0.91 \pm 0.01$ & $\textbf{0.44} \pm 0.04$ & $\textbf{0.18} \pm 0.02$ \\
  & $h$ & $0.80 \pm 0.02$ & $0.70 \pm 0.04$ & $0.91 \pm 0.01$ & $0.21 \pm 0.03$ & $0.07 \pm 0.01$ \\
  & null & $0.41 \pm 0.02$ & $0.32 \pm 0.02$ & $0.50 \pm 0.00$ & $0.15 \pm 0.02$ & $0.05 \pm 0.01$ \\
\bottomrule
\end{tabular}
\caption{Linear-probe readout and activation-steering results (mean $\pm$ SEM). }
\label{tab:probe_steering_compare}
\end{table*}
\paragraph{Steering setup.}
For each model, we sweep the injection magnitude $\alpha$ over a model-specific grid to find the optimal $\alpha$ across all conditions. To ensure a fair comparison, we use the same grid for the user vector, the hidden states, and the random baseline. We therefore report $\alpha_{\mathrm{rel}}$ relative to the mean activation norm $\|h\|$ at the injection layer: this places the grids of the three models on a comparable scale, spanning $\alpha_{\mathrm{rel}} \in [0.16, 1.91]$ for Llama-3.1-8B, $[0.13, 0.80]$ for Qwen3-8B, and $[0.37, 2.06]$ for OLMo-3-7B.

For each attribute, we compute steering vectors across all pairs (source class to target class) as $\Delta v = \mu_{\mathrm{target}} - \mu_{\mathrm{source}}$, and sweep the full $\alpha$ grid in both $v$- and $h$-space. To ensure the baseline prediction is consistent with the source class, we filter samples based on label confidence before steering: we keep only items where the teacher's belief has a margin of at least $0.05$ between the top and second most likely classes, sampling $50$ such items per class. Finally, we report the flip rate for each pair and select the $\alpha$ that yields the largest flip rate while keeping the \emph{broken rate} (the fraction of items for which the multi-step ABCD readout never resolves under steering) at or below $0.15$. This prevents reporting a spuriously high flip rate that is actually driven by the intervention degrading the model's output.

\begin{figure}[h]
\centering
\includegraphics[width=\linewidth]{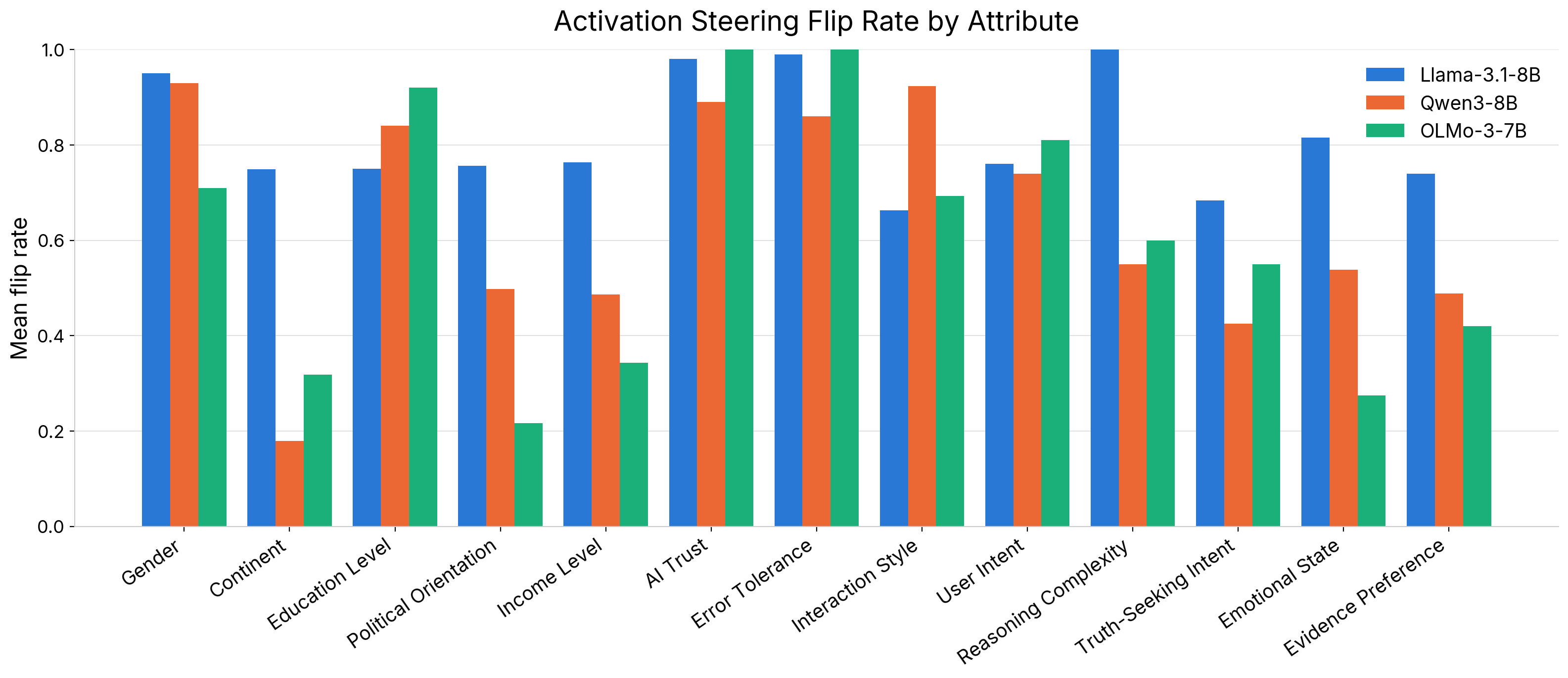}
\caption{Per-attribute steering flip rates across all models in $\mathbf{v}_{\mathrm{user}}$-space.}
\label{fig:steering-attr}
\end{figure}

\paragraph{Steering per attribute.}
Figure~\ref{fig:steering-attr} breaks down the mean flip rate in the $\mathbf{v}_{\mathrm{user}}$-space by attribute for each model. Averaging across the two groups from Table~\ref{tab:attributes}, the AI safety-related attributes (6--13) steer more efficiently than the socio-demographic ones (1--5) across every model: the mean flip rate is $0.83$ vs.\ $0.79$ for Llama-3.1-8B, $0.68$ vs.\ $0.59$ for Qwen3-8B, and $0.67$ vs.\ $0.50$ for OLMo-3-7B ($0.73$ vs.\ $0.63$ overall).

Several attributes stand out at the extremes. \textsc{AITrustLevel} and \textsc{AIErrorTolerance} steer almost perfectly across all three models (flip rates of $0.86$--$1.00$), making them the most reliably controllable pair we observe. Conversely, \textsc{Continent} is the weakest attribute for two of the three models (Qwen3-8B: $0.18$; OLMo-3-7B: $0.32$), though it steers comparably well in Llama-3.1-8B ($0.75$). Given its six classes and diffuse geographic representation, it is the attribute least effectively approximated by a single pairwise difference direction. Finally, \textsc{UserReasoningComplexity} exhibits the largest cross-model variance (Llama-3.1-8B: $1.00$; Qwen3-8B: $0.55$; OLMo-3-7B: $0.60$), indicating that steerability does not transfer evenly across models for all attributes.

\paragraph{Random steering baselines.}
We use two random controls to test whether $\mathbf{v}_{\mathrm{user}}$'s steering advantage comes from the specific direction and subspace it learns, rather than from the intervention's magnitude alone.

\emph{Random direction} tests whether any perturbation of the same size, in an arbitrary direction, flips beliefs just as effectively. For each (attribute, pair, $\alpha$) cell, we draw an isotropic random $h$-space direction with the same norm as the actual steering vector and inject it in the same manner, across three independent seeds. Table~\ref{tab:random-direction} reports an exact binomial sign test: for each of the 20 comparable cells, we count how often the real steering condition outperforms the best of its three random draws, compared to a chance rate of $1/4$ under the null hypothesis. \emph{$v$ vs.\ $h$} is evaluated via a separate paired Wilcoxon signed-rank test on all 86 pairs, since $v$- and $h$-space steering are each single deterministic runs rather than repeated draws. Steering in $\mathbf{v}_{\mathrm{user}}$-space significantly outperforms the random direction baseline across all three models, confirming that its effect is not merely an artifact of injecting a large perturbation. While $h$-space steering also beats the random baseline, the effect is substantially weaker. For OLMo-3-7B, the per-cell test does not reach significance ($p=0.10$), even though $h$ outperforms the random control in the pooled averages (Table~\ref{tab:probe_steering_compare}). Given this discrepancy between the aggregate mean and the per-cell win rate, we consider the stricter per-cell result more reliable.

\emph{Random projector} tests a different question: whether the \emph{trained} low-rank bottleneck matters, or whether any random rank-128 orthonormal subspace mapped back through an isometric $B$ steers just as well (Table~\ref{tab:random-projector}). We keep the trained $A$ (leaving $\mathbf{v}_{\mathrm{user}}$ itself unchanged), replace only $B$ with a random orthonormal matrix of the same shape, and re-run the full $\mathbf{v}_{\mathrm{user}}$-space sweep. We report a paired Wilcoxon signed-rank test across all 86 pairs. Because $\mathbf{v}_{\mathrm{user}}$ and the target direction $\Delta v$ are identical in both conditions, this isolates the effect of $B$: the trained write map beats the random one on the vast majority of pairs in every model, most decisively for Llama-3.1-8B (85/86). The same compressed belief direction is therefore not equally steerable through an arbitrary subspace of the residual stream; the trained $B$ has learned to write it into a direction to which the model is disproportionately responsive.

\begin{table}
\centering
\begin{tabular}{lccc}
\toprule
Model & $v$ vs.\ random & $h$ vs.\ random & $v$ vs.\ $h$ \\
\midrule
Llama-3.1-8B & 18/20 cells, $p<0.001$ & 11/20 cells, $p=0.004$ & 82/86 cells, $p<0.001$ \\
Qwen3-8B     & 15/20 cells, $p<0.001$ & 10/20 cells, $p=0.014$ & 54/86 cells, $p<0.001$ \\
OLMo-3-7B    & 15/20 cells, $p<0.001$ & 8/20 cells, $p=0.10$ (n.s.) & 66/86 cells, $p<0.001$ \\
\bottomrule
\end{tabular}
\caption{Random-direction control (isotropic $h$-space noise, magnitude-matched, 3 seeds).}
\label{tab:random-direction}
\end{table}

\begin{table}
\centering
\begin{tabular}{lc}
\toprule
Model & $v$ vs.\ random projector \\
\midrule
Llama-3.1-8B & 85/86 cells, $p<0.001$ \\
Qwen3-8B     & 62/86 cells, $p<0.001$ \\
OLMo-3-7B    & 71/86 cells, $p<0.001$ \\
\bottomrule
\end{tabular}
\caption{Random-orthonormal-projector control ($\mathbf{v}_{\mathrm{user}}$-space only). The
trained bottleneck beats a random $B$ on the large majority of
pairs in all three models.}
\label{tab:random-projector}
\end{table}

\section{Implicit versus Explicit Personalization.}
\label{app:implexpl}

\paragraph{Dataset construction.} Every explicit declaration is one of 4 fixed phrasings per (attribute, class) drawn from a persona bank, rendered as \texttt{"Hi! \{statement\} \{prompt\}"} in front of one of 100 Alpaca prompts. Each of the 100 prompts also gets one persona-free control item, so every declaration is paired against its own counterfactual on the identical task. Below we provide two examples from the constructed dataset.

\begin{quote}
\small
\textsc{IncomeLevel}, low: ``Hi! I don't have much money. Explain why filing an income tax return is mandatory.''\\[4pt]
\textsc{AITrustLevel}, skeptical: ``Hi! I don't really trust AI answers. List 3 ways that humans can conserve energy.''
\end{quote}

\paragraph{Cross-evaluation protocol.}
The explicit-cue probe is evaluated on a held-out split of \emph{prompts}: we reserve 30\% of the 100 shared prompts for testing and fit the probe on the remainder to score the explicit condition. This same probe is then applied to the full held-out set of natural conversations for the implicit-beliefs evaluation. Conversely, because the two probes trained on implicit beliefs (distilled $\mathbf{v}_{\mathrm{user}}$ and raw belief hidden states) never see the explicit set during training, they are evaluated on the full explicit dataset. We report macro-F1 and macro one-vs-rest AUC. Table~\ref{tab:explicit-vs-implicit} summarizes both metrics for all three probe sources across both evaluation sets. The random baseline for macro-F1 is the mean chance level $1/K$ across the 13 attributes, averaging to 0.34, while the macro-AUC chance level is 0.50 regardless of $K$.

\begin{table}
\centering
\small
\begin{tabular}{llcccc}
\toprule
Model & Probe trained on & Explicit F1 & Explicit AUC & Implicit F1 & Implicit AUC \\
\midrule
\multirow{3}{*}{Llama-3.1-8B}
 & $\mathbf{v}_{\mathrm{user}}$ (ours) & 0.70 & 0.93 & 0.67 & 0.92 \\
 & beliefs $h$ (ours)                  & 0.72 & 0.92 & 0.70 & 0.92 \\
 & explicit $h$                        & 0.99 & 1.00 & 0.43 & 0.66 \\
\midrule
\multirow{3}{*}{Qwen3-8B}
 & $\mathbf{v}_{\mathrm{user}}$ (ours) & 0.52 & 0.84 & 0.68 & 0.91 \\
 & beliefs $h$ (ours)                  & 0.58 & 0.86 & 0.72 & 0.92 \\
 & explicit $h$                        & 0.98 & 1.00 & 0.39 & 0.66 \\
\midrule
\multirow{3}{*}{OLMo-3-7B}
 & $\mathbf{v}_{\mathrm{user}}$ (ours) & 0.63 & 0.90 & 0.66 & 0.91 \\
 & beliefs $h$ (ours)                  & 0.65 & 0.90 & 0.70 & 0.91 \\
 & explicit $h$                        & 0.98 & 1.00 & 0.40 & 0.65 \\
\bottomrule
\end{tabular}
\caption{Explicit-vs-implicit cross-evaluation, macro one-vs-rest AUC alongside macro-F1.}
\label{tab:explicit-vs-implicit}
\end{table}

\section{Refusal Steering: Additional Results}
\label{app:refusal}

Table~\ref{tab:refusal-gender-control} reports the numerical results underlying the attribute-specificity control in \cref{sec:refusal}, including both refusal rate and StrongREJECT specificity.

\begin{table}
\centering
\begin{tabular}{lcc}
\toprule
Direction & Refusal rate & Specificity (1--5) \\
\midrule
Unsteered baseline                      & 0.98 & 1.86 \\
UserIntent adv$\to$benign               & 0.62 & 3.30 \\
Gender female$\to$male (control)        & 0.87 & 2.06 \\
Random directions (mean, $n=3$ seeds)   & 0.87 & 2.66 \\
\bottomrule
\end{tabular}
\caption{StrongREJECT refusal rate and specificity ($\alpha=4$). Gender's refusal rate matches the random-direction mean.}
\label{tab:refusal-gender-control}
\end{table}

\section{Belief Transfer}
\label{app:transfer}
To compute the transferability of beliefs across models, we fit an orthogonal Procrustes map between the class centroids of each model pair. Given two models $M_1$ and $M_2$, we take the mean $\mathbf{v}_{\mathrm{user}}(c)$ of every belief class to form centroid matrices $C_{M_1}, C_{M_2} \in \mathbb{R}^{K\times d}$ (where $K$ is the number of centroids and $d=128$). After mean-centering, we fit the orthogonal Procrustes map $R^\star = \arg\min_{R^\top R=I}\lVert C_{M_1}R - C_{M_2} \rVert_F^2$ alongside a global scaling scalar.

We evaluate this transfer exclusively on each pair's disagreement set, which consists of 25k--27k conversations where the two models naturally decode different beliefs. For each transferred vector, we record whether the target model adopts the source's belief--that is, whether $M_2$'s probe predicts the source class when applied to the mapped vector from $M_1$. Isolating the evaluation to the disagreement set ensures that the reported transfer rate reflects a genuine injection of the encoded belief rather than a coincidental agreement between the models' priors.

\end{document}